%% file: main.tex
\documentclass[conference]{IEEEtran}
\IEEEoverridecommandlockouts
\usepackage{cite}
\usepackage{amsmath,amssymb,amsfonts}
\usepackage{algorithmic}
\usepackage{graphicx}
\usepackage{textcomp}
\usepackage{xcolor}
\def\BibTeX{{\rm B\kern-.05em{\sc i\kern-.025em b}\kern-.08em
    T\kern-.1667em\lower.7ex\hbox{E}\kern-.125emX}}

\makeatletter
\newcommand{\linebreakand}{%
  \end{@IEEEauthorhalign}
  \hfill\mbox{}\par
  \mbox{}\hfill\begin{@IEEEauthorhalign}
}
\makeatother

\usepackage[lined,boxed,commentsnumbered,ruled]{algorithm2e}
\usepackage{algorithmic}

\newcommand{\hlt}[1]{#1}

\newcommand{\nop}[1]{}
\usepackage{xcolor}
\usepackage{color, soul}

\newtheorem{definition}{\textbf{Definition}}

\DeclareMathOperator*{\argmax}{arg\,max}

\usepackage{amsmath,mathtools,bm,etoolbox}
\usepackage{bbm}

\DeclarePairedDelimiterXPP\Aver[1]{\mathbb{E}}{[}{]}{}{

#1
}

\begin{document}

\title{Revealing Unfair Models by Mining Interpretable Evidence
%
}

\author{\IEEEauthorblockN{Mohit Bajaj}
\IEEEauthorblockA{\textit{Huawei Technologies Canada} \\
Burnaby, Canada \\
mohit.bajaj1@huawei.com}
\and
\IEEEauthorblockN{Lingyang Chu}
\IEEEauthorblockA{\textit{McMaster University} \\
Hamilton, Canada \\
chul9@mcmaster.ca}
\and
\IEEEauthorblockN{Vittorio Romaniello}
\IEEEauthorblockA{\textit{University of British Columbia} \\
Vancouver, Canada \\
vittorio.romaniello@stat.ubc.ca}
\and
\IEEEauthorblockN{Gursimran Singh}
\IEEEauthorblockA{\textit{Huawei Technologies Canada} \\
Burnaby, Canada\\
gursimran.singh1@huawei.com}
\linebreakand
\IEEEauthorblockN{Jian Pei}
\IEEEauthorblockA{\textit{Simon Fraser University} \\
Burnaby, Canada\\
jpei@cs.sfu.ca}
\and
\IEEEauthorblockN{Zirui Zhou}
\IEEEauthorblockA{\textit{Huawei Technologies Canada} \\
Burnaby, Canada\\
zirui.zhou@huawei.com}
\and
\IEEEauthorblockN{Lanjun Wang}
\IEEEauthorblockA{\textit{Tianjin University} \\
Tianjin, China \\
wanglanjun@tju.edu.cn}
\and
\IEEEauthorblockN{Yong Zhang}
\IEEEauthorblockA{\textit{Huawei Technologies Canada} \\
Burnaby, Canada\\
yong.zhang3@huawei.com}
}


\maketitle

\begin{abstract}
The popularity of machine learning has increased the risk of unfair models getting deployed in high-stake applications, such as justice system, drug/vaccination design, and medical diagnosis.
Although there are effective methods to train fair models from scratch, how to automatically reveal and explain the unfairness of a trained model remains a challenging task.
Revealing unfairness of machine learning models in interpretable fashion is a critical step towards fair and trustworthy AI.
In this paper, we systematically tackle the novel task of \textit{revealing unfair models by mining interpretable evidence (RUMIE)}. 
The key idea is to find solid evidence in the form of a group of data instances discriminated most by the model.
To make the evidence interpretable, we also find a set of human-understandable key attributes and decision rules that characterize the discriminated data instances and distinguish them from the other non-discriminated data.
As demonstrated by extensive experiments on many real-world data sets, our method finds highly interpretable and solid evidence to effectively reveal the unfairness of trained models. Moreover, it is much more scalable than all of the baseline methods. 
\end{abstract}


\input{sections/1_Introduction}

\input{sections/2_RelatedWork_old}

\input{sections/3_ProblemFormulation}

\input{sections/4_Relaxing}

\input{sections/5_Solution}
\input{sections/6_Experiments}

\input{sections/7_Conclusion}

\bibliographystyle{IEEETran}
\bibliography{acmart}

\end{document}

%% file: sections/1_Introduction.tex

\section{Introduction}
\label{sec:intro}

Deployment of unfair machine learning models in the society, if unchecked, can lead to significant discrimination against certain groups of subjects~\cite{berk2012criminal, ahmad2018interpretable, vamathevan2019applications, galindo2000credit}.
For example, a particular group of prisoners may get discriminated by an AI-assisted justice system to receive unfair judgements~\cite{propub}. A group of consumers may be discriminated by not allowing them to avail some premium services based on their communities and neighborhoods~\cite{amazon}.

\nop{
and a minority group of patients may be discriminated and ignored by an automated drug design model~\cite{xxx}.
}

Many fair machine learning methods~\cite{kamiran2012data, chakraborty2021bias, kamishima2012fairness, kamiran2018exploiting, lawless2021fair} focus on training fair models, however, there has been little control on the fairness of models trained by third-parties~\cite{propub}.
Therefore, before deploying a trained model, it is crucial to reveal and explain potential unfairness of the model.


\emph{How can we reveal and explain the unfairness of a trained model on a given data distribution?}

Existing model explanation methods~\cite{carvalho2019machine} cannot reveal and explain the unfairness of a trained model, because they focus on answering the questions of why and how a machine learning model makes a prediction, but they ignore the fairness of the prediction.

We attempt to tackle this challenging task by borrowing the principled idea of statistical hypothesis testing~\cite{shi2008statistical}. Specifically, we assume that the model is fair and then try to reject the assumption by finding the most extreme evidence in data to demonstrate the unfairness of the model.

Intuitively, the most extreme evidence in data is the group of data instances that are discriminated the most by the model. 
A group of data instances is said to be discriminated if an instance in the group has a much lower probability of getting a favorable prediction, such as mortgage approval and college entrance admission, than the other data instances not contained in the group \cite{dwork2012fairness}.

To explain why a group of data instances gets discriminated by the model, it is also important to identify the set of key attributes that characterizes the group and distinguishes the instances in the group from the rest of the data instances..
\nop{
The group of discriminated data instances are often characterized by multiple key attributes which distinguishes them from the other data instances that are not in the group.
It is also important to identify the key attributes, because they are essential to interpreting why the group of data instances gets discriminated by the machine learning model.
}
For example, to demonstrate that a correctional offender management system is unfair, one can find solid evidence in data, such as the group of African-American with dark skin and are younger than 25 who are subject to a much higher false negative rate on releasing decisions than the other suspects~\cite{report}. 
The key attribute values ``African-American'' and ``dark skin'' explain why the group of suspects is discriminated by the system.

Following the above intuition, we propose a novel data mining task named \textbf{Revealing Unfair Models by mining Interpretable Evidence (RUMIE)}, which aims to reveal  and explain potential unfairness of a trained model by finding the most discriminated group of instances and the corresponding key attributes.

RUMIE is a challenging task for multiple reasons. 
First, the discriminated group of data instances is often characterized by multiple key attributes. Thus, the time cost of searching the discriminated group and the key attributes grows exponentially with respect to the number of attributes. Continuous attributes make this task further challenging as there may exist huge number of unique assignments for such attributes.
Second, it is challenging to interpret the discriminated group and key attributes to an end user without machine learning background, because the user may not understand how and why the key attributes characterize the discriminated group.


\nop{
unfairness of a loan-approval system, we can find solid evidence in data by identifying a group of people who are getting an extremely lower approval rate than the other people. Presumably, 

the group of unmarried female not owning a house who are subject to a much lower approval rate of loan than the other people 

is a solid evidence to demonstrate the unfairness of a banking system. This group of female 

the group of unmarried female not owning a house may be discriminated by a banking system, thus are subject to a much lower approval rate of loan than the other people. 
}

\nop{
An intuitive and effective way is to find \textbf{interpretable evidence} in data, that is, the group of data instances that are the most discriminated by the machine learning model. 
}

\nop{
Here, the discriminated group is characterized by three attributes: “unmarried”, “female”, and “not owning a house”.

More often than not, data instances in a discriminated group are characterized by multiple attributes which distinguishes the discriminated data instances from the other data instances not contained in the discriminated group. Identifying such attributes is essential to interpret why a discriminated group gets discriminated by a machine learning model.
}

\nop{
To interpret why a discriminated group of data instances are discriminated, it is essential to identify the set of attributes that characterizes the 
discriminated group of data instances and distinguishes them from the other data instances not in the group.

For example : 
The group of unmarried females not owning a house is subject to a biased low probability to successfully getting a loan from a bank.
Here, the group is defined by three attributes: “unmarried”, “female”, and “not owning a house”.
}


In this paper, we make the following contributions. 

First, we propose the novel task of RUMIE. The goal is to find a discriminated group of instances that is most discriminated by a trained model. We also aim to interpret why the group of instances is discriminated by finding a small set of key attributes that characterizes the discriminated group of instances and distinguishes them from the other data instances that are not discriminated.

Second, we propose an effective algorithm to efficiently find the most discriminated group of data instances. While the scalability of brute-force searching methods is significantly limited by the exponential number of combinations of attributes, our method achieves outstanding efficiency by formulating a continuous numerical optimization problem, which can be efficiently solved by a well designed method.

Third, to explain to the users how the key attributes characterize the discriminated group,
we use the key attributes to construct a decision tree that separates the discriminated group from the other data instances. Then, we derive human-understandable rules from the decision tree to explain how the key attributes distinguish the discriminated group from the other data instances.
As shown in extensive case studies, the key attributes and the decision tree provides clear interpretation on why the group of instances are discriminated.

Last, we report extensive experiments on many real-world data sets to demonstrate the superior performance and scalability of our method.

%% file: sections/2_RelatedWork_old.tex

\section{Related work}
\label{sec:related}

In this section, we discuss the relationship between our work and four category of related works, such as discrimination discovery, software testing, fair model training and fairness evaluation tools.

\subsection{Discrimination Discovery and Software Testing}
Discrimination discovery approaches are related to our work because discrimination often indicates unfairness. Thus, methods developed to discover the discrimination in data are potentially related to revealing the unfairness of a machine learning model.

Many discrimination discovery methods have been developed to find discrimination in databases. 
Please see \cite{romei2014multidisciplinary} for an excellent survey. Some of these works focus on individual discrimination\cite{luong2011k,zhang2016situation} where the goal is to identify individual instances in the data sets that are discriminated by the model in contrast to other similar instances that are favored by the model. 
There also exist some methods that align with our goal of revealing unfairness in models by identifying groups of instances being discriminated. Among these works, our work is the most related to the following.
The DDD method~\cite{ruggieri2010data} uses frequent item set mining to find the most common assignments of the categorical attributes of data instances and then analyze discrimination based on the common assignments.
FairDB~\cite{azzalini2021fair} uses Approximate Conditional Functional Dependencies (ACFDs) to capture discriminating association rules between attribute assignments and predictions. 

Both the methods use pattern mining techniques to discover the discrimination in databases. 
Since the adopted pattern mining techniques cannot effectively deal with real-valued data, these methods cannot effectively find evidence in real-valued data to reveal the unfairness of a trained machine learning model.

It is possible to extended DDD~\cite{ruggieri2010data} and FairDB~\cite{azzalini2021fair} to tackle the RUMIE task by quantizing real-valued data attributes into categorical ones. 
However, as demonstrated by extensive experiments in Section~\ref{sec:performance}, both the methods cannot achieve good performance because due to the information loss caused by the quantization.


Another related work is a software testing method named Themis~\cite{galhotra2017fairness}, which was originally proposed to test the unfairness of a software system in a data-independent manner.
The key idea is to build test cases by adaptively sampling synthetic data from the uniform distribution of all possible combinations of data attributes.

The correctness of Themis relies on the adaptive confidence-driven sampling strategy, which is difficult to extend to sampling from real data.
As discussed later in Section~\ref{sec:exp}, Themis can be extended to find a solution to the RUMIE task.
However, the solution cannot effectively reveal the unfairness of a model on a set of real data instances, because the results produced by Themis are based on synthetic data instances, which are independent from real data.

\subsection{Fair Model Training}

The family of fair-model-training methods~\cite{kamiran2012data, chakraborty2021bias, kamishima2012fairness, kamiran2018exploiting, lawless2021fair} focus on mitigating the bias of models.
The pre-processing methods~\cite{kamiran2012data, chakraborty2021bias} attempt to mitigate the bias of a model by modifying the training data with pre-processing techniques, such as removing and masking sensitive attributes like ``gender'', ``race'', ``age'', etc.
The in-processing methods~\cite{kamishima2012fairness, calders2010three, goh2016satisfying} mitigate the bias of a model by smoothly incorporating model fairness in the training objective of the model. 
The post-processing methods~\cite{kamiran2018exploiting} focus on reversing the biased predictions made by an unfair trained model. 

The fair-model-training methods are effective in training fair models from scratch.
However, they cannot be straight-forwardly extended to tackle the RUMIE task because they do not find the most discriminated group of instances to reveal the unfairness of a trained model.


%
%

Some methods~\cite{berk2021fairness, o2016weapons, eubanks2018automating, chouldechova2018frontiers, friedler2019comparative, verma2018fairness, binns2018s, kleinberg2016inherent, zafar2017fairness} develop new measures of fairness and leverage such measures to train fair models.
Demographic parity~\cite{dwork2012fairness} computes the fractions of data instances predicted positive in multiple groups and takes the absolute value of the difference of the fractions to measure discrimination.
Equality of false positive rate~\cite{hardt2016equality} computes the fraction of data instances falsely predicted positive in the true negative population of each group. It improves model fairness by requiring such fractions of all groups to be the same.
Equality of false discovery rate~\cite{kleinberg2016inherent} requires the percentage of false positive predictions in all positive predictions to be the same for each group of data instances. 

The above methods focus on developing fairness measures to train fair models, but they do not find evidence to reveal the unfairness of a trained model. Therefore, they cannot be directly applied to tackle the RUMIE task.

\subsection{Fairness Evaluation Tools}

There are also several published tools~\cite{bellamy2019ai, wexler2019if, udeshi2018automated} to compute the unfairness of trained models, such as the AI Fairness 360 Open Source Toolkit~\cite{bellamy2019ai},
the What-If Tool~\cite{wexler2019if}, and the open source bias audit toolkit named Aequitas ~\cite{udeshi2018automated}.

These interactive tools are developed to compute existing fairness metrics when the group of data instances and sensitive attributes are provided.
They rely on users to input a group of data instances or a set of sensitive attributes, and then output the significance of unfairness of the model with respect to the user inputs.

These tools cannot tackle the RUMIE task because they serve substantially different purposes. For the RUMIE task, the discriminated group of data instances and the corresponding key attributes are not known beforehand. The goal of RUMIE is to automatically find such groups of data instances and attributes to reveal the unfairness of a trained model.








%% file: sections/3_ProblemFormulation.tex

\section{Problem formulation}
\label{sec:pf}

In this section, we first introduce how to evaluate the significance of discrimination for a group of data instances, then we define the RUMIE task and formulate it as a continuous optimization problem.




%

\subsection{Evaluating the Significance of Discrimination}

Discrimination and fairness are like two sides of the same coin.
A low level of fairness usually means a high level of discrimination.
Therefore, we can measure the significance of discrimination of a machine learning model based on the fairness measurements of the model~\cite{feldman2015certifying, calders2009building, dwork2012fairness, calders2010three, zhang2020fairfl, hardt2016equality}.

Existing model fairness measurements can be categorized into the following two major categories. 

The individual fairness measurements~\cite{dwork2012fairness, joseph2016fairness} evaluate whether a data instance is treated fairly by a model by investigating how likely a pair of similar instances receive similar predictions from the model. 
A data instance is said to be treated fairly by the model if it receives a similar prediction with that of another similar data instance.

The group fairness measurements~\cite{feldman2015certifying, 
calders2009building,  
calders2010three,
hardt2016equality,
donini2018empirical} evaluate the fairness of a model on all data instances by evaluating how balanced the predictions of the model are on a pre-defined pair of protected groups of data instaces (e.g., the group of `female' versus the group of `male'). 

Since our goal is to reveal the unfairness of a trained machine learning model, we are more interested in measuring the fairness of the model on all data instances instead of individual data instances. 
Therefore, the group fairness measurements are more suitable for our task.


Many group fairness measurements have been developed in literature, such as disparate impact~\cite{feldman2015certifying}, demographic parity~\cite{calders2009building, dwork2012fairness, calders2010three}, difference of F1-scores~\cite{zhang2020fairfl}, equal odds~\cite{hardt2016equality}, and equal opportunities~\cite{hardt2016equality}.
Many of these measurements can be incorporated into our proposed framework to reveal the unfairness of models.
Without loss of generality, we focus on developing our framework based on demographic parity~\cite{calders2009building, dwork2012fairness, calders2010three}. 
In the end of Section~\ref{sec:rumietask}, 
we will discuss the key idea to incorporate some other group fairness measurements into our framework.


Next, we give a brief introduction to demographic parity~\cite{calders2009building, dwork2012fairness, calders2010three} and extend it to a discrimination score to measure the discrimination significance of a model on a group of data instances.


Denote by $g$ a trained classification model and by $\mathcal{D}$ a set of data instances. 
Among all the classes, we assume that there is one \textbf{favorable class/prediction}.  The choice of favorable prediction is application dependent. For example, it may be getting a mortgage application approved in a bank or a suspect being freed by a law system. We denote by $\mathtt{fav} $ the favorable class/prediction.

Denote by $\mathcal{S}\subset\mathcal{D}$ a group of data instances in $\mathcal{D}$ and by $\mathcal{D}\setminus\mathcal{S}$ the other group containing the rest of the data instances in $\mathcal{D}$.
Let $\mathbb{P}(\mathtt{fav}  \mid \mathcal{S})$ and $\mathbb{P}(\mathtt{fav}  \mid \mathcal{D} \setminus \mathcal{S})$ be the probabilities of a data instance $\mathbf{x}\in\mathcal{D}$ receiving a favorable prediction from $g$ when $\mathbf{x}$ is uniformly sampled from $\mathcal{S}$ and $\mathcal{D}\setminus \mathcal{S}$, respectively.

According to the definition of demographic parity~\cite{calders2009building, dwork2012fairness, calders2010three}, the model $g$ is said to be fair with respect to the groups $\mathcal{S}$ and $\mathcal{D}\setminus\mathcal{S}$ if
\begin{equation}
\mathbb{P}(\mathtt{fav}  \mid \mathcal{S}) = \mathbb{P}(\mathtt{fav}  \mid \mathcal{D} \setminus \mathcal{S}),
\end{equation}
which means the instances in the groups $\mathcal{S}$ and $\mathcal{D}\setminus \mathcal{S}$ have equal probabilities of receiving favorable predictions.

Demographic parity is a boolean measurement that only tells whether a model is fair or not. It cannot be directly applied to measure the discrimination significance of a model.
Therefore, we extend demographic parity to compute the \textbf{discrimination score} of $g$ on $\mathcal{S}$ as
\begin{equation}\label{eq:discrimination_score}
	DScore(\mathcal{S}) = \mathbb{P}(\mathtt{fav}  \mid \mathcal{D} \setminus \mathcal{S})-\mathbb{P}(\mathtt{fav}  \mid \mathcal{S}), 
\end{equation}
which has a real valued range from $-1$ to $1$ to measure the significance of discrimination of a model.

If $DScore(\mathcal{S})=0$, then $\mathbb{P}(\mathtt{fav}  \mid \mathcal{S}) = \mathbb{P}(\mathtt{fav}  \mid \mathcal{D} \setminus \mathcal{S})$, thus the model $g$ is fair.

If $DScore(\mathcal{S})\in[-1, 0)$, then $\mathbb{P}(\mathtt{fav}  \mid \mathcal{S}) > \mathbb{P}(\mathtt{fav}  \mid \mathcal{D} \setminus \mathcal{S})$, which means a data instance in $\mathcal{S}$ has a higher probability to receive a favorable prediction than a data instance in $\mathcal{D}\setminus\mathcal{S}$.
Therefore, the group $\mathcal{S}$ is not discriminated by the model $g$.

If $DScore(\mathcal{S})\in(0, 1]$, then $\mathbb{P}(\mathtt{fav}  \mid \mathcal{S}) < \mathbb{P}(\mathtt{fav}  \mid \mathcal{D} \setminus \mathcal{S})$, which means a data instance in $\mathcal{S}$ has a lower probability to receive a favorable prediction than a data instance in $\mathcal{D}\setminus\mathcal{S}$. Therefore, the group $\mathcal{S}$ is discriminated by the model $g$.

A larger value of $DScore(\mathcal{S})$ means that an instance in $\mathcal{S}$
has a lower probability to receive a favorable prediction than an instance in $\mathcal{D}\setminus\mathcal{S}$, which further indicates a more significant discrimination on $\mathcal{S}$.



\subsection{The RUMIE Task}
\label{sec:rumietask}

In this subsection, we first define the RUMIE task. Then we propose a probabilistic framework to model the RUMIE task as a continuous optimization problem. Last, we discuss the key idea to incorporate some other group fairness measurements into our framework.

\begin{definition}[RUMIE task]
Given a classification model $g$, a set of data instances $\mathcal{D}$, and a set of sensitive attributes $\mathcal{A}$ that may be potentially related to the discrimination of $g$ on groups of instances in $\mathcal{D}$, the task of \textbf{revealing unfair models by mining interpretable evidence (RUMIE)} is to
\begin{enumerate}
    \item find the \textbf{group of data instances}, denoted by $\mathcal{S}\subset\mathcal{D}$, that is the most discriminated by the model $g$; and 
    \item interpret the discrimination on the data instances in $\mathcal{S}$ by finding a small set of \textbf{key attributes}, denoted by $\mathcal{Q}\subseteq \mathcal{A}$, such that the data instances in $\mathcal{S}$ are separated from those in $\mathcal{D} \setminus \mathcal{S}$ in the feature space of $\mathcal{Q}$.
\end{enumerate}
\end{definition}

Here, the feature space of $\mathcal{Q}$, denoted by $\mathtt{Space}^\mathcal{Q}$, is the space where a data instance is represented by the attributes in $\mathcal{Q}$. 
The discriminated group $\mathcal{S}$ is said to be separated from $\mathcal{D}\setminus\mathcal{S}$ in $\mathtt{Space}^\mathcal{Q}$ if $\mathcal{S}$ and $\mathcal{D}\setminus\mathcal{S}$ are classified as two different classes by the classification boundary of a binary classifier in $\mathtt{Space}^\mathcal{Q}$.  

To clearly explain to an end user why and how $\mathcal{S}$ is separated from $\mathcal{D}\setminus\mathcal{S}$, 
we also want the classification boundary separating $\mathcal{S}$ and $\mathcal{D}\setminus\mathcal{S}$ to be simple.
Here, a classification boundary is said to be simple if the partitions of data set $\mathcal{D}$ induced by the classification boundary have a small minimum description length.
For example, a linear classification boundary of a logistic regression classifier is said to be simpler than a non-linear classification boundary of a deep neural network.


We refer to $(\mathcal{S}, \mathcal{Q})$ as the \textbf{interpretable evidence} to demonstrate the unfairness of $g$.
The discriminated group $\mathcal{S}$ is the evidence to demonstrate the unfairness of $g$ because $\mathcal{S}$ is discriminated the most by $g$.
A higher discrimination score on $\mathcal{S}$ means more unfairness exists in $g$.
The set of key attributes $\mathcal{Q}$ allows us to interpret the reason behind the discrimination of $\mathcal{S}$ because these attributes are used to separate the discriminated group $\mathcal{S}$ from the other group $\mathcal{D}\setminus\mathcal{S}$. A simpler classification boundary to separate $\mathcal{S}$ and $\mathcal{D}\setminus\mathcal{S}$ in $\mathtt{Space}^\mathcal{Q}$ indicates  the existence of the discriminated group is less possible by chance, and thus is more significant.

Tackling the RUMIE task requires to
find the optimal set of $\mathcal{S}$ that maximizes the discrimination score $DScore(\mathcal{S})$; and it also requires to find a small set of sensitive attributes denoted by $\mathcal{Q}$ such that $\mathcal{S}$ and $\mathcal{D}\setminus\mathcal{S}$ are well separated by a simple decision boundary in $\mathtt{Space}^\mathcal{Q}$.
This can be done by a brute-force search approach that enumerates all values of $\mathcal{S}$ and $\mathcal{Q}$ to find their optimal values.
However, the brute-force search approach does not scale up well on large data sets, because the time cost grows exponentially with respect to the size of $\mathcal{D}$ and $\mathcal{A}$.

%% file: sections/4_Relaxing.tex
To efficiently tackle the RUMIE task, we propose a probabilistic framework to model it as a continuous optimization problem, which can be efficiently solved by a gradient-based optimization method~\cite{lillo1993solving}.

The key idea is to model the discrete variables $\mathcal{S}$ and $\mathcal{Q}$ by functions of continuous variables, and then formulate the optimization objective of RUMIE as a continuous function.

\textbf{First}, we model $\mathcal{S}$ as a random set of data instances, where the probability of a data instance $\mathbf{x}\in\mathcal{D}$ belonging to $\mathcal{S}$ is modelled by
\begin{align}\label{eq:prob_member}
    \mathbb{P}(\mathbf{x}\in\mathcal{S}) = f_{\boldsymbol\theta}(\sigma_\mathcal{A}(\mathbf{x})).
\end{align}
Here, $\sigma_\mathcal{A}(\mathbf{x})$ is the projection of $\mathbf{x}$ to the feature space $\mathtt{Space}^\mathcal{A}$ of the attributes in $\mathcal{A}$, $f_{\boldsymbol\theta}: \mathtt{Space}^\mathcal{A} \rightarrow [0,1]$ is a continuous function parametrized by a vector $\boldsymbol\theta$ with ${|\mathcal{A}|}$ entries, and $|\mathcal{A}|$ is the number of attributes in $\mathcal{A}$. Every entry in $\boldsymbol\theta$ is a continuous real-valued variable.




For a given set of data instances $\mathcal{D}$, the distribution of $\mathcal{S}$ is determined by $\boldsymbol\theta$. 
Therefore, $\mathcal{S}$ is a function of the continuous variables in $\boldsymbol\theta$.

\textbf{Second}, we model $\mathcal{Q}$ as a function of continuous variables by formulating $f_{\boldsymbol\theta}$ as a logistic regression classifier, that is, 
 \begin{align}\label{eq:classifier}
	f_{\boldsymbol\theta}(\sigma_\mathcal{A}(\mathbf{x})) = \frac{1}{1+e^{-\boldsymbol\theta^\top \sigma_\mathcal{A}(\mathbf{x})}},
\end{align}
where the none-zero entries in $\boldsymbol\theta$ correspond to the set of key attributes $\mathcal{Q}\in\mathcal{A}$ because the data instances in  $\mathcal{S}$ and $\mathcal{D}\setminus\mathcal{S}$ are well separated by the simple linear decision boundary induced by $f_{\boldsymbol\theta}$ in $\mathtt{Space}^\mathcal{Q}$.
Since the value of $\mathcal{Q}$ is determined by $\boldsymbol\theta$, $\mathcal{Q}$ is a function of continuous variables.

As illustrated later, we will enhance the interpretability of $\mathcal{Q}$ by reducing the size of $\mathcal{Q}$ with a sparsity constraint on ${\boldsymbol\theta}$ and generate a set of more interpretable rules by distilling $f_{\boldsymbol\theta}$ into a simple decision tree.

%




%

\textbf{Third}, we introduce how to formulate the optimization objective of the RUMIE task as a continuous function of $\boldsymbol\theta$.

Since the RUMIE task requires to find the group of data instances that is the most discriminated by the model $g$, we need to find a discriminated group $\mathcal{S}$ that maximizes the discrimination score $DScore(\mathcal{S})$ in Equation~\eqref{eq:discrimination_score}.

Since we model $\mathcal{S}$ as a random set of data instances parametrized by $\boldsymbol\theta$, the discrimination score $DScore(\mathcal{S})$ is a real-valued random variable, which cannot be straight-forwardly maximized.
Therefore, instead of directly maximizing the random variable $DScore(\mathcal{S})$, we seek to maximize its expectation
\begin{equation}
\begin{aligned}
	\Aver{DScore(\mathcal{S})} = \Aver{\mathbb{P}(\mathtt{fav} \mid \mathcal{D} \setminus \mathcal{S})} - \Aver{\mathbb{P}(\mathtt{fav} \mid \mathcal{S})},
\end{aligned}
\end{equation}
where $\Aver{\mathbb{P}(\mathtt{fav} \mid \mathcal{D} \setminus \mathcal{S})}$ and $\Aver{\mathbb{P}(\mathtt{fav} \mid \mathcal{S})}$ are the expectations of $\mathbb{P}(\mathtt{fav} \mid \mathcal{S})$ and $\mathbb{P}(\mathtt{fav} \mid \mathcal{D} \setminus \mathcal{S})$, respectively.
Here, $\mathbb{P}(\mathtt{fav} \mid \mathcal{S})$ and $\mathbb{P}(\mathtt{fav} \mid \mathcal{D} \setminus \mathcal{S})$ are random variables because $\mathcal{S}$ is a random set of data instances.

Recall that $\mathbb{P}(\mathtt{fav} \mid \mathcal{S})$ is the probability of a data instance $\mathbf{x}$ receiving a favorable prediction from $g$ when it is uniformly sampled from $\mathcal{S}$. We write $\mathbb{P}(\mathtt{fav} \mid \mathcal{S})$ as
\begin{align}\label{eq:pfav1}
\mathbb{P}(\mathtt{fav} \mid \mathcal{S}) = 
\frac{\sum_{\mathbf{x}\in\mathcal{D}} \mathbbm{1}_\mathtt{fav}(\mathbf{x})*\mathbbm{1}(\mathbf{x}\in\mathcal{S})}
{\sum_{\mathbf{x}\in\mathcal{D}} \mathbbm{1}(\mathbf{x}\in\mathcal{S})}.
\end{align}
where $\mathbbm{1}_\mathtt{fav}(\mathbf{x})\in\{0,1\}$ is an indicator function that is equal to 1 only if $g$ makes a favorable prediction on $\mathbf{x}$; and $\mathbbm{1}(\mathbf{x}\in\mathcal{S})\in\{0,1\}$ is an indicator function that is equal to 1 only if $\mathbf{x}\in\mathcal{S}$.

Since every data instance $\mathbf{x}$ has a probability of $\mathbb{P}(\mathbf{x}\in\mathcal{S})$ to belong to $\mathcal{S}$, the indicator function
$\mathbbm{1}(\mathbf{x}\in\mathcal{S})$ is a random variable following a Bernoulli distribution, and its expectation is
\begin{align}
\Aver{\mathbbm{1}(\mathbf{x}\in\mathcal{S})} = \mathbb{P}(\mathbf{x}\in\mathcal{S}).
\end{align}
Since the numerator and denominator in Equation~\eqref{eq:pfav1} are conditionally independent given ${\boldsymbol\theta}$, we have 
\begin{align}
\Aver{\mathbb{P}(\mathtt{fav} \mid \mathcal{S})} 
& = \frac{\sum_{\mathbf{x}\in\mathcal{D}} \mathbbm{1}_\mathtt{fav}(\mathbf{x})*\Aver{\mathbbm{1}(\mathbf{x}\in\mathcal{S})}}
{\sum_{\mathbf{x}\in\mathcal{D}} \Aver{\mathbbm{1}(\mathbf{x}\in\mathcal{S})}}\\
& = \frac{\sum_{\mathbf{x}\in\mathcal{D}} \mathbbm{1}_\mathtt{fav}(\mathbf{x})*\mathbb{P}(\mathbf{x}\in\mathcal{S})}
{\sum_{\mathbf{x}\in\mathcal{D}} \mathbb{P}(\mathbf{x}\in\mathcal{S})}.
\end{align}
Similarly, we can derive
\begin{align}
\Aver{\mathbb{P}(\mathtt{fav} \mid \mathcal{D}\setminus\mathcal{S})} = 
\frac{\sum_{\mathbf{x}\in\mathcal{D}} \mathbbm{1}_\mathtt{fav}(\mathbf{x})*\mathbb{P}(\mathbf{x}\in\mathcal{D}\setminus\mathcal{S})}
{\sum_{\mathbf{x}\in\mathcal{D}} \mathbb{P}(\mathbf{x}\in\mathcal{D}\setminus\mathcal{S})},
\end{align}
where $\mathbb{P}(\mathbf{x}\in\mathcal{D}\setminus\mathcal{S})=1-\mathbb{P}(\mathbf{x}\in\mathcal{S})$ is the probability of an instance $\mathbf{x}$ not belonging to $\mathcal{S}$.

Recall that $\mathbb{P}(\mathbf{x}\in\mathcal{S})=f_{\boldsymbol\theta}(\sigma_\mathcal{A}(\mathbf{x}))$ is a function of $\boldsymbol\theta$. Thus both $\Aver{\mathbb{P}(\mathtt{fav} \mid \mathcal{S})}$ and $\Aver{\mathbb{P}(\mathtt{fav} \mid \mathcal{D}\setminus\mathcal{S})}$ are real-valued functions of $\boldsymbol\theta$. Therefore, $\Aver{DScore(\mathcal{S})}$ is a rea-valued function of $\boldsymbol\theta$, which can be maximized with respect to $\boldsymbol\theta$.

\textbf{Last}, we model the RUMIE task as the following \textbf{continuous optimization problem}, which aims to maximize the expected discrimination score on $\mathcal{S}$.
\begin{equation}\label{eq:ContOptim}
	\begin{aligned}
		\argmax_{\boldsymbol\theta} \; & \Aver*{DScore(\mathcal{S})} - \lambda * \mathcal{C}(k, \boldsymbol\theta), \\
		&\; \textrm{s.t.}\; \alpha \leq \frac{\Aver{|\mathcal{S}|}}{|\mathcal{D}|} \leq \beta,
	\end{aligned}
\end{equation}
where $\lambda > 0$, $\alpha\in[0,1]$ and $\beta\in[0,1]$, $\alpha\leq\beta$, are positive real-valued hyper-parameters, $\Aver{|\mathcal{S}|}$ is the expected size of $\mathcal{S}$ and $|\mathcal{D}|$ is the size of $\mathcal{D}$.

The second term $\lambda * \mathcal{C}(k, \boldsymbol\theta)$ in the objective of the optimization problem is a regularization term that reduces the size of $\mathcal{Q}$ to enhance the interpretability of the discrimination evidence. Here $k$ refers to size of $\mathcal{Q}$ and implies the maximum number of sensitive attributes to be considered for distinguishing $\mathcal{S}$ from $\mathcal{D}\setminus\mathcal{S}$. Choosing $k$ provides the user with the flexibility to adjust the interpretability as per their needs. 
The term
\begin{align}\label{eq:k_constr}
    \mathcal{C}(k, \boldsymbol\theta) = \frac{1}{|\mathcal{A}|-k}\sum_{j=1}^{|\mathcal{A}|-k}smallest(\boldsymbol\theta,j),
\end{align}
is the L-1 reduced penalization term adopted from~\cite{zhang2010analysis, ahn2017difference,huang2015nonconvex} . The function $smallest(\boldsymbol\theta, j)$ returns the $j$-th smallest absolute value of the entries in ${\boldsymbol\theta}$.



Maximizing the objective minimizes the term $\mathcal{C}(k, \boldsymbol\theta)$, which enforces the vector ${\boldsymbol\theta}$ to contain $k$ effective non-zero entries while squeezing the rest of the entries to be very small or zero~\cite{huang2015nonconvex, ahn2017difference}. 
After finding a solution $\boldsymbol\theta^*$ to the continuous optimization problem, we strictly enforce the maximum limit of $k$ sensitive attributes by keeping only the $k$ entries with the largest absolute values in $\boldsymbol\theta^*$ and setting the other entries in $\boldsymbol\theta^*$ to zero.

Comparing to using the L1-norm of $\boldsymbol\theta$ as a regularization term to induce the sparsity of $\boldsymbol\theta$, $\mathcal{C}(k, \boldsymbol\theta)$ provides a better control on the number of non-zero entries in $\boldsymbol\theta$~\cite{zhang2010analysis, ahn2017difference,huang2015nonconvex}.
A small value of $k$ leads to a small set of $\mathcal{Q}$ thus produces a clear interpretation for the reason behind discrimination of a group.

In the constraint of the optimization problem,
$\alpha$ and $\beta$ constrain the expected proportion taken by $\mathcal{S}$ in $\mathcal{D}$. Since different end users may be interested in different sizes of discriminated instance groups for different applications, the thresholds $\alpha$ and $\beta$ provide a good flexibility to suit the needs of different end users. 
For example, a bank's fairness policy may only be violated if at least 20\% of its clients are being discriminated by a deployed AI lender model.


Next, we discuss how to incorporate some other group fairness measurements, such as disparate impact~\cite{feldman2015certifying}, difference of F1-scores~\cite{zhang2020fairfl}, equal odds~\cite{hardt2016equality}, and equal opportunities~\cite{hardt2016equality} into the proposed framework.

The key idea is to extend a fairness measurement into a discrimination score $DScore(\mathcal{S})$, and then compute the expected discrimination score $\Aver{DScore(\mathcal{S})}$ on the distribution of $\mathcal{S}$. 
Since the distribution of $\mathcal{S}$ is parameterized by $\boldsymbol\theta$, $\Aver{DScore(\mathcal{S})}$ is a continuous function of $\boldsymbol\theta$, and it can be directly plugged in Equation~\eqref{eq:ContOptim} to formulate a continuous optimization problem for the RUMIE task.

It is straight-forward to extend a fairness measurement to a discrimination score. 
The real-valued measurements such as difference of F1-scores~\cite{zhang2020fairfl} and disparate impact~\cite{feldman2015certifying} can be directly used as discrimination scores and can be optimized with respect to $\boldsymbol\theta$.
The boolean measurements such as equal odds~\cite{hardt2016equality}, and equal opportunities~\cite{hardt2016equality} can be extended to a real-valued discrimination score in a similar way as we extend demographic parity~\cite{calders2009building, dwork2012fairness, calders2010three}. 
Deriving the expectation of a discriminated score on the distribution of $\mathcal{S}$ is straight-forward, thus we skip the details.

%% file: sections/5_Solution.tex
 \section{Mining Interpretable Evidences}
 
In this section, we first introduce how to find the set of key attributes $\mathcal{Q}$ and the discriminated group $\mathcal{S}$
by solving the continuous optimization problem. 
Then, we introduce a heuristic method to trade the discrimination score on $\mathcal{S}$ for a better interpretation on $\mathcal{S}$.

\subsection{Finding the Key Attributes and the Discriminated Group}

To find the set of key attributes $\mathcal{Q}$ and the discriminated group $\mathcal{S}$, 
we first find a local optimal solution $\boldsymbol\theta^*$ to the continuous optimization problem in Equation~\eqref{eq:ContOptim}, then we derive $\mathcal{Q}$ and $\mathcal{S}$ from $\boldsymbol\theta^*$.

The continuous optimization problem is a standard constrained continuous optimization problem, which can be effectively and efficiently solved by the classic penalty method~\cite{lillo1993solving, van2015penalty, yeniay2005penalty}. 
The solution $\boldsymbol\theta^*$ is a local optimal solution because the loss function is non-convex.

We use stochastic gradient descent (SGD)~\cite{bottou2010large} to optimize $\boldsymbol\theta$. The time complexity of SGD is linear with respect to the number of attributes in $\mathcal{A}$, thus our method can find a local optimal solution to the RUMIE task much faster than the other baselines that exhaustively search the feature space $\mathtt{Space}^\mathcal{A}$ of the attributes in $\mathcal{A}$. Also our method's runtime is insensitive to choice of $k$ where as the runtime of other methods increases substantially with increase in $k$ due to exponential increase in search space of possible attribute combinations.

Next, we introduce how to derive the set of key attributes $\mathcal{Q}$ from the solution $\boldsymbol\theta^*$. 

Recall that $f_{\boldsymbol\theta^*}$ is the classifier to separate $\mathcal{S}$ from $\mathcal{D}\setminus\mathcal{S}$, the attributes corresponding to the top-$k$ largest absolute values in $\boldsymbol\theta^*$ are taken as the set of key attributes $\mathcal{Q}$.
To keep $\boldsymbol\theta^*$ consistent with $\mathcal{Q}$, we also update $\boldsymbol\theta^*$ by reserving only those $k$ entries corresponding to the key attributes in $\mathcal{Q}$ and setting the other entries to zero.

Deriving the discrimination group $\mathcal{S}$ from $\boldsymbol\theta^*$ is straightforward.
Since the function $f_{\boldsymbol\theta^*}$ is a logistic regression classifier that separates the data instances in $\mathcal{S}$ and $\mathcal{D}\setminus\mathcal{S}$, that is,
\begin{align}
    \mathbf{x}\in\left\{
    \begin{aligned}
   	&\mathcal{S} \; & if\; f_{\boldsymbol\theta^*}(\sigma_\mathcal{A}(\mathbf{x}))\geq 0.5,\\
    	&\mathcal{D}\setminus\mathcal{S} \; & if\; f_{\boldsymbol\theta^*}(\sigma_\mathcal{A}(\mathbf{x})) < 0.5.
    \end{aligned}
    \right.
\end{align}
Therefore, we can derive the discriminated group as $\mathcal{S}=\{\mathbf{x}\in\mathcal{D} \mid f_{\boldsymbol\theta^*}(\sigma_\mathcal{A}(\mathbf{x}))\geq 0.5\}$.

Algorithm~\ref{alg:find} summarizes how to compute the discriminated group $\mathcal{S}$, the set of key attributes $\mathcal{Q}$, and the solution $\boldsymbol\theta^*$ to the continuous optimization problem.
Steps 2-3 finds a local optimal solution $\boldsymbol\theta^*$ to the continuous optimization problem;
steps 4-8 derives the set of attributes $\mathcal{Q}$ from $\boldsymbol\theta^*$; and steps 9-12 derives the group $\mathcal{S}$ from $\boldsymbol\theta^*$.

\hlt{
Since $\boldsymbol\theta^*$ is a local optimal solution to the continuous optimization problem, 
the discriminated group $\mathcal{S}$ returned by Algorithm~\ref{alg:find} may not have the largest discrimination score, and there may exist other discriminated groups with larger discrimination scores.
However, this does not compromise our goal to reveal the unfairness of the model $g$, 
because the existence of the discriminated group $\mathcal{S}$ returned by Algorithm~\ref{alg:find} is already a solid evidence to reveal that the model $g$ is at least as unfair as indicated by the discrimination score on $\mathcal{S}$.
}

\hlt{
As demonstrated by extensive experiments in Section~\ref{sec:exp}, solving the continuous optimization problem can effectively and efficiently find solid evidence to reveal the unfairness of the model $g$, and the discriminated group derived from this solution corresponds to much more significant discrimination than those found by the other strong baseline methods.
}

%
%
%
%
%
%


\begin{algorithm}[t]
\small
\caption{Find Key Attributes and Discriminated Group}
\label{alg:find}
	\KwIn{A trained model $g$, a set of data instances $\mathcal{D}$, and a set of sensitive attribute $\mathcal{A}$.}
	\KwOut{The set of key attributes $\mathcal{Q}$, the discriminated group $\mathcal{S}$, and the solution $\boldsymbol\theta^*$.}
	\begin{algorithmic}[1]
	
	\STATE Initialization: $\mathcal{Q}\leftarrow\emptyset$ and $\mathcal{S}\leftarrow\emptyset$.
	
	\STATE Solve the continuous optimization problem in Equation~\eqref{eq:ContOptim} \\by penalty method~\cite{lillo1993solving} to get a solution $\boldsymbol\theta^*$.
	
	\STATE Update $\boldsymbol\theta^*$ by keeping only the top-$k$ entries with the largest \\absolute value and set the rest of the entries to zero.
	
	\FOR{each sensitive attribute $a\in\mathcal{A}$}
		\IF{The corresponding entry of $a$ in $\boldsymbol\theta^*$ is not zero}
			\STATE $\mathcal{Q}\leftarrow \mathcal{Q}\cup a$.
		\ENDIF
	\ENDFOR
	
	\FOR{each data instance $\mathbf{x}\in\mathcal{D}$}
		\IF{$f_{\boldsymbol\theta^*}(\sigma_\mathcal{A}(\mathbf{x})) \geq 0.5$}
			\STATE $\mathcal{S}\leftarrow\mathcal{S}\cup\mathbf{x}$.
		\ENDIF
	\ENDFOR
	
	\RETURN $\mathcal{Q}$, $\mathcal{S}$, and $\boldsymbol\theta^*$.
	\end{algorithmic}
\end{algorithm}

\nop{
can sometimes be too mathematical to be easily understood by an end-user.

These evidence are easy to be understood 
However, these evidences can sometimes be too mathematical to be easily understood by an end-user.

Next, we introduce how to enhance the interpretability of the by training a highly interpretable decision tree.
}

\subsection{Trading Discrimination Score for Better Interpretability}

For an end user without machine learning background, the discriminated group $\mathcal{S}$ and the set of key attributes $\mathcal{Q}$ found by Algorithm~\ref{alg:find} are not easy to understand because it is still unclear how the key attributes in $\mathcal{Q}$ work to separate $\mathcal{S}$ from $\mathcal{D}\setminus\mathcal{S}$.



In the example of the correctional offender management system~\cite{report}, a discriminated group $\mathcal{S}$ is a group of suspects and the set $\mathcal{Q}$ consists of three attributes, ``race'', ``skin color'' and ``age''.
Since each of the three attributes can take many different values, simply showing the attributes to an end user without telling them specific combinations of attribute values that are associated with the discrimination cannot effectively explain how the group of discriminated people are separated from the non-discriminated group.
To effectively explain to end users, we need to produce human-understandable rules, such as ``the discriminated group is \textit{African-American} with \textit{dark skin} and \textit{younger than 25}'', 
where it is clear how specific values of ``race'', ``skin color'' and ``age'' combine to characterize the discriminated group.
Moreover, a discriminated group can be large and may contain several sub-groups of suspects, such as 
``\textit{African-American} with \textit{dark skin} and \textit{younger than 25}'',
``\textit{Asian-American} with \textit{dark skin}'', 
and ``\textit{Asian-American} with \textit{light skin} and \textit{older than 30}''.
An effective and intuitive approach is to properly organize these sub-groups in a decision tree to clearly explain them to users.


Next, we introduce how to translate the evidence $(\mathcal{S},\mathcal{Q})$ into human-understandable rules.
The key idea is to train a decision tree to separate $\mathcal{S}$ from $\mathcal{D}\setminus\mathcal{S}$ in $\mathtt{Space}^\mathcal{Q}$. 
In this way, the decision rules generated by the decision tree interpret how the key attributes in $\mathcal{Q}$ work to separate $\mathcal{S}$ from $\mathcal{D}\setminus\mathcal{S}$.




A naive method is to directly train a decision tree, denoted by $\mathcal{T}$, to separate $\mathcal{S}$ from $\mathcal{D}\setminus\mathcal{S}$ in $\mathtt{Space}^\mathcal{Q}$ such that $\mathcal{S}$ is the set of positive instances classified by $\mathcal{T}$.
However, since the piece-wise linear decision boundary of a decision tree is substantially different from the linear decision boundary of the logistic regression classifier in Equation~\eqref{eq:classifier}, $\mathcal{T}$ can be a complicated tree with large depth.
In this case, the decision rules of $\mathcal{T}$ may be too complicated to understand.
In order to generate simple rules that are easy to understand, we have to prune $\mathcal{T}$ to produce a shallow decision tree $\mathcal{P}$ whose depth is as small as possible. 


Denote by $\mathcal{S}'\subseteq\mathcal{D}$ the set of discriminated instances classified by $\mathcal{P}$.
Due to the information loss caused by pruning, $\mathcal{S}'$ may not be exactly the same as $\mathcal{S}$, which means the discrimination score of $\mathcal{S}'$ may not be as large as that of $\mathcal{S}$.
However, $\mathcal{S}'$ can still be a good evidence to reveal the unfairness of the model $g$ if its discrimination score is not significantly less than that of $\mathcal{S}$.

Recall that the decision rules of $\mathcal{P}$ will be more interpretable than that of $\mathcal{T}$.
Thus using $\mathcal{S}'$ as the evidence to the RUMIE task is essentially trading discrimination score for better interpretability.







Based on the above insights, we formulate the following decision-tree-based translation problem
to trade discrimination score for better interpretability.

\begin{definition}[Decision-Tree-Based Translation problem]
Given a decision tree $\mathcal{T}$ trained to separate $\mathcal{S}$ and $\mathcal{D}\setminus\mathcal{S}$ in $\mathtt{Space}^\mathcal{Q}$, the \textbf{decision-tree-based translation problem} is to prune $\mathcal{T}$ into a shallow decision tree $\mathcal{P}$ such that
\begin{enumerate}
	\item the depth of $\mathcal{P}$ is as small as possible; and
	\item the groups $\mathcal{S}'$ and $\mathcal{D}\setminus\mathcal{S}'$ classified by $\mathcal{P}$ satisfy the size constraints of RUMIE task mentioned in Equation~\eqref{eq:ContOptim}.
\end{enumerate}
\end{definition}


%
%

Algorithm~\ref{alg:enhance} summarizes our method to tackle this problem.
We adopt the minimal cost-complexity pruning (MCP) algorithm~\cite{breiman2017classification,salzberg1994c4} to prune $\mathcal{T}$. This algorithm has a parameter $\psi \geq 0$ to control the level of pruning. 
For a larger value of $\psi$, $\mathcal{T}$ will be pruned more heavily, thus $\mathcal{P}$ will be a shallower tree with a smaller depth. 

\hlt{
The key idea of Algorithm~\ref{alg:enhance} is to find the largest value of $\psi$ such that the size constraints on $\mathcal{S}'$ and $\mathcal{D}\setminus\mathcal{S}'$ mentioned in Equation~\eqref{eq:ContOptim} are still satisfied.}
Then, we return $\mathcal{S}'$ and $\mathcal{D}\setminus\mathcal{S}'$ as the interpretable evidence, and use the decision rules derived from $\mathcal{P}$ to explain how the key attributes distinguish $\mathcal{S}'$ from $\mathcal{D}\setminus\mathcal{S}'$.

Extensive case studies in Section~\ref{sec:case} demonstrate the outstanding interpretability of the evidences found by our method. In Section~\ref{sec:performance}, we also demonstrate that this interpretability comes with very little loss in discrimination score corresponding to $\mathcal{S}$.

\begin{algorithm}[t]
\small
\caption{Decision-tree-based Translation}
\label{alg:enhance}
	\KwIn{A set of data instances $\mathcal{D}$, the discriminated group $\mathcal{S}$, the set of key attributes $\mathcal{Q}$.}
	\KwOut{A shallow decision tree $\mathcal{P}$.}
	\begin{algorithmic}[1]
	
	
	\STATE Train a decision tree $\mathcal{T}$ to separate $\mathcal{S}$ from $\mathcal{D}\setminus\mathcal{S}$ in $\mathtt{Space}^\mathcal{Q}$.
	
	\STATE Set the parameter $\psi$ of the (MCP) algorithm~\cite{salzberg1994c4} to zero.
	
	\REPEAT
		\STATE $\psi\leftarrow\psi+\epsilon$, where $\epsilon > 0$ is a small fixed step-size.
		\STATE Prune $\mathcal{T}$ by MCP~\cite{salzberg1994c4} to produce a pruned decision tree $\mathcal{P}$.
		\STATE Find $\mathcal{S}'\subseteq \mathcal{D}$ that is classified by $\mathcal{P}$ as the discriminated group.
	\UNTIL{$\mathcal{S}'$ or $\mathcal{D}\setminus\mathcal{S}'$ no longer satisfy size constraints of Equation~\eqref{eq:ContOptim}}
	\RETURN The last $\mathcal{P}$ when $\mathcal{S}'$ and $\mathcal{D}\setminus\mathcal{S}'$ satisfied the size constraints of Equation~\eqref{eq:ContOptim}.
	\end{algorithmic}
\end{algorithm}

%% file: sections/6_Experiments.tex
\section{Experiments}
\label{sec:exp}



In this section, we introduce the baseline methods and the evaluation metrics that we use for comparison. 
Then, we introduce the data sets and discuss the setup of the RUMIE task on each data set.
Next, we conduct interesting case studies to comprehensively demonstrate the outstanding interpretability of the evidences found by our method.
Last, we quantitatively analyze the experimental performance of all compared methods and discuss why our method can achieve superior performance than the other baseline methods.

\begin{figure*}[t]
    \begin{center}
       	\includegraphics[width=1\linewidth]{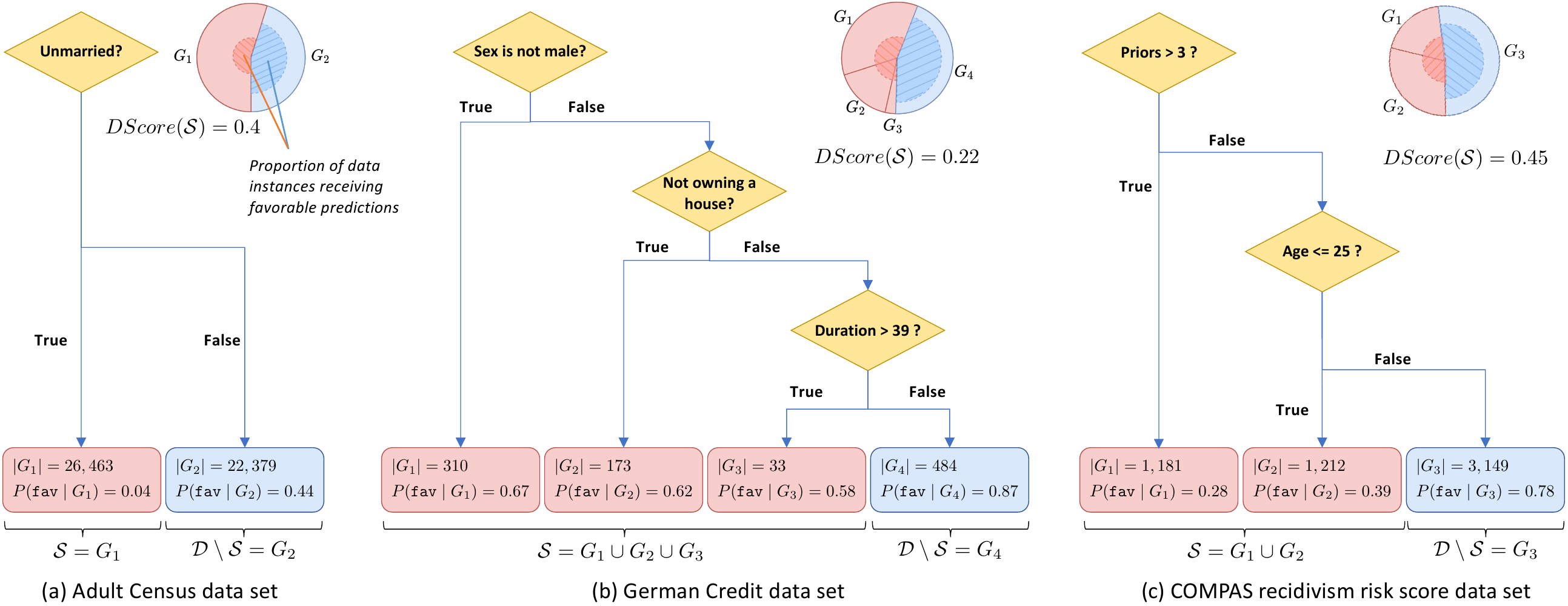}
    \end{center}
\caption{The decision trees produced by IE-DT on each of the data sets. The set of data instances in a leaf node is denoted by $\{G_1, G_2, \ldots\}$. A leaf node in red means the set of data instances belongs to the discriminated group $\mathcal{S}$. A leaf node in blue means the set of data instances belong to the not-discriminated group $\mathcal{D}\setminus\mathcal{S}$. 
For each tree, we draw a pie chart to show the proportion of the data instances contained in each leaf node. The red and blue shaded areas within the pie chart shows the proportions of data instances in $\mathcal{S}$ and $\mathcal{D}\setminus\mathcal{S}$, respectively, that receive favorable predictions from the model $g$.
}
\label{fig:case_study}
\end{figure*}

\subsection{Baseline Methods and Evaluation Metrics}
As discussed previously in Section~\ref{sec:related}, the RUMIE task is a novel task that has not been addressed systematically. 
The closest baseline methods are as follows.

The first baseline that we design is named \textbf{Enum}. It is an enumeration method that exhaustively searches all the combinations of all values of the sensitive attributes in $\mathcal{A}$ such that maximum number of attributes considered in the combinations are not more than $k$. For each continuous variable, it selects one of the unique values of the variable as a threshold and discretizes the rest values based on if they are greater than the selected threshold or not. This is repeated for each unique value of the continuous attribute where it is chosen as the threshold and rest of the values are discretized based on it.   
This baseline is effective in finding good quality solutions, however, the time cost grows exponentially with respect to the number of attributes.




The second baseline is \textbf{Themis}~\cite{galhotra2017fairness}. As introduced in Section~\ref{sec:related}, Themis was originally proposed to test the discrimination of a software system based on sampled synthetic data instances.
The output of Themis is the values of a subset of sensitive attributes, which define a group of synthetic data instances discriminated by the tested software system.


To extend Themis for the RUMIE task, we first use Themis to test the model $g$ as the software system and get the output of Themis, that is, the set of values for maximum of $k$ number of sensitive attributes. 
Then, we regard the subset of sensitive attributes as $\mathcal{Q}$ and obtain $\mathcal{S}$ by extracting the data instances in $\mathcal{D}$ whose values of attributes in $\mathcal{Q}$ match the values returned by Themis.
For the comprehensiveness of our experiment, we compare with 3 versions of Themis, namely, Themis-1k, Themis-500, Themis-100, where the numbers 1k, 500, and 100 represent the numbers of sampled synthetic data instances drawn from each possible combination of attributes.

The third baseline we use is DDD~\cite{ruggieri2010data} that uses frequent itemset mining to find common attribute assignments to reveal significant discrimination patterns. We adapt it to RUMIE task by returning the assignment that satisfies the constraints of RUMIE task and achieves maximum discrimination score. 

The last baseline to our method is FairDB~\cite{azzalini2021fair} that uses Approximate Conditional Functional Dependencies (ACFDs) to detect unfairness. This method also returns list of assignments that are associated with significant discrimination. We take the assignment with maximum discrimination value satisfying the constraints of RUMIE task as output of this method. 

Themis, DDD and FairDB are not originally designed to process continuous attributes. We adopt the three bin quantization method used by Themis~\cite{galhotra2017fairness} to quantize the continuous attributes into categorical ones for these methods.

We use two versions of our method in the experiments.
The first one named \textbf{Interpretable Evidence (IE)} corresponds to Algorithm~\ref{alg:find}  and the second one named \textbf{Interpretable Evidence - Decision Tree (IE-DT)} corresponds to Algorithm~\ref{alg:enhance}. The continuous optimization formulated in Equation~\eqref{eq:ContOptim} naturally addresses continuous attributes, thus no quantization is required for IE and IE-DT.


The performance of all the methods is evaluated using two criteria : scalability and solution quality.
Scalability is measured by observing how the \textbf{runtime} (i.e., cost of time) to find the evidence changes with increase in value of $k$ and total number of sensitive attributes. 
Solution quality is measured using the second metric is the \textbf{discrimination score} $DScore(\mathcal{S})$.
For each method, a higher discrimination score means a more significant discrimination on the discriminated group $\mathcal{S}$, which further indicates a better effectiveness in finding solid evidence to reveal model unfairness.

The code of Themis~\cite{galhotra2017fairness}, DDD~\cite{ruggieri2010data} and FairDB~\cite{azzalini2021fair} is published by the respective authors. The code for Enum, IE, and IE-DT is available at~\cite{our_code}.
For all of the methods, we carefully tune the parameters and report the best performance in our experiments.


%
%

\subsection{Data Sets and Task Setup}
We use the following three widely-adopted public data sets in our experiments.

The Adult Census data set~\cite{kohavi1996scaling} was extracted from the database of year 1994 in the Census Bureau of USA.
This data set contains the census information of 48,842 adults.
Every adult is represented by a data instance that consists of 94 categorical attributes and 6 continuous attributes.
For each adult, a predictive model is used to predict whether they make more than \$50,000 a year or not. The favorable prediction is making more than \$50,000 a year.

The German Credit data set~\cite{Dua:2019} contains the information of 1,000 individuals who take credit from a bank.
Every individual is represented by a data instance consisting of 17 categorical attributes and 3 continuous attributes. 
The credit risk of every person is classified as high or low by a predictive model. The favorable prediction is a low credit risk.

The COMPAS recidivism risk score data set~\cite{propub} contains the information of 6,172 individuals from Broward County in Florida of the U.S. 
Every individual is represented by a data instance that consists of 9 categorical attributes and 1 continuous attribute. 
For each individual, the risk of recidivism in two years is predicted by the predictive model named COMPAS~\cite{brennan2009evaluating}.
The favorable prediction in this data set is low recidivism risk.

For each of the above data sets, we set up the RUMIE task by regarding the predictive model as the model $g$, and the entire set of data instances as $\mathcal{D}$.
Deciding which attributes among all of the attributes to consider sensitive and include in $\mathcal{A}$ often depends on task-related prior knowledge and ethics code of the concerned organizations. 
Without loss of generality, we use the entire set of attributes as $\mathcal{A}$.

The outcome of the RUMIE task 
on the Adult Census data set is interpretable evidence to demonstrate severe discrimination made by the model when predicting annual incomes of certain group of individuals. 
On the German Credit data set, the outcome will be solid interpretable evidence revealing the unfairness of the predictive model in predicting credit risks. 
For the COMPAS data set, we find interpretable evidence to demonstrate the unfairness of the COMPAS model in judging the risk of recidivism of prisoners.

\subsection{Case Studies}
\label{sec:case}

In this section, we present some interesting case studies to demonstrate the outstanding interpretability and quality of the evidence found by IE-DT.
Figure~\ref{fig:case_study} shows the decision tree $\mathcal{P}$ computed by IE-DT on each of the data sets. The parameters in Equation~\eqref{eq:ContOptim} are set to $k=5$, $\alpha=0.45$, $\beta=0.55$, and $\lambda=1$ for all data sets.
The influence of parameters will be discussed later in Section~\ref{sec:performance}.

Figure~\ref{fig:case_study}(a) shows the decision tree computed on the Adult Census data set, where IE-DT successfully finds a significantly discriminated group $\mathcal{S}$ with a large $DScore(\mathcal{S})=0.4$. 

\textit{Who are the people in $\mathcal{S}$ and why are they discriminated?} The decision tree $\mathcal{P}$ in Figure~\ref{fig:case_study}(a) provides a clear explanation.
The discriminated group $\mathcal{S}$ represented by $G_1$ is the group of unmarried people, and the not-discriminated group $\mathcal{D}\setminus\mathcal{S}$ represented by $G_2$ is the group of married people.
A married person in $G_2$ has a probability of 0.44 to receive a favorable prediction, however, for an unmarried person in $G_1$, this probability drops significantly to 0.04.
Now we know the unmarried people are significantly discriminated by the predictive model, which is solid evidence to reveal the unfairness of the model. 

It is also interesting to see in Figure~\ref{fig:case_study}(a) that the discriminated group $\mathcal{S}$ of unmarried people contains more than half of the people in the data set. Therefore, a large population subgroup is being significantly discriminated and may be a matter of concern.

%



\begin{figure*}[h!]
    \begin{center}
        \includegraphics[width=1.0\linewidth]{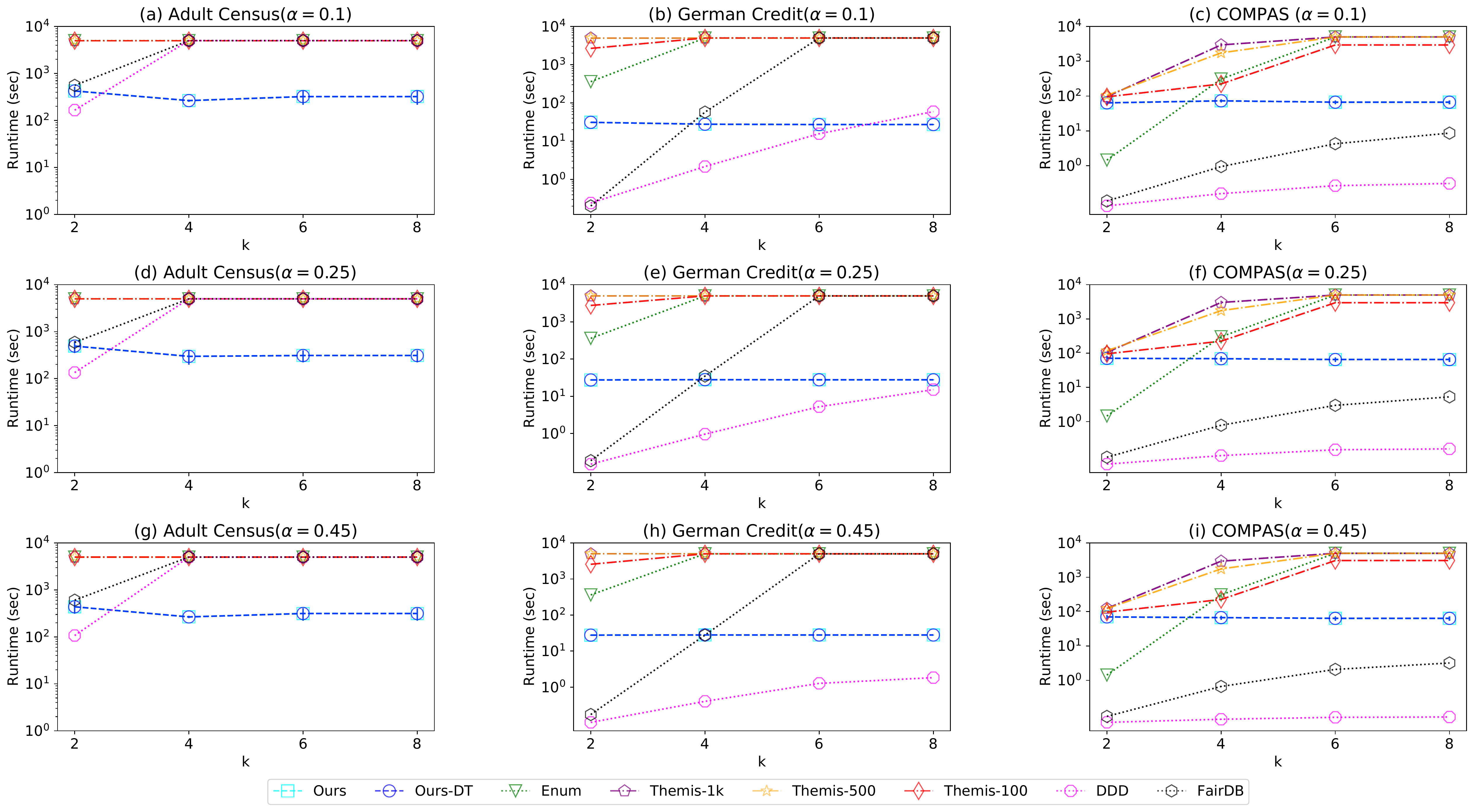}

    \end{center}
\caption{The runtime performance of all compared methods on each of the data sets.}
\label{fig:runtime}
\end{figure*}

\begin{figure*}[h!]
    \begin{center}
        \includegraphics[width=1.0\linewidth]{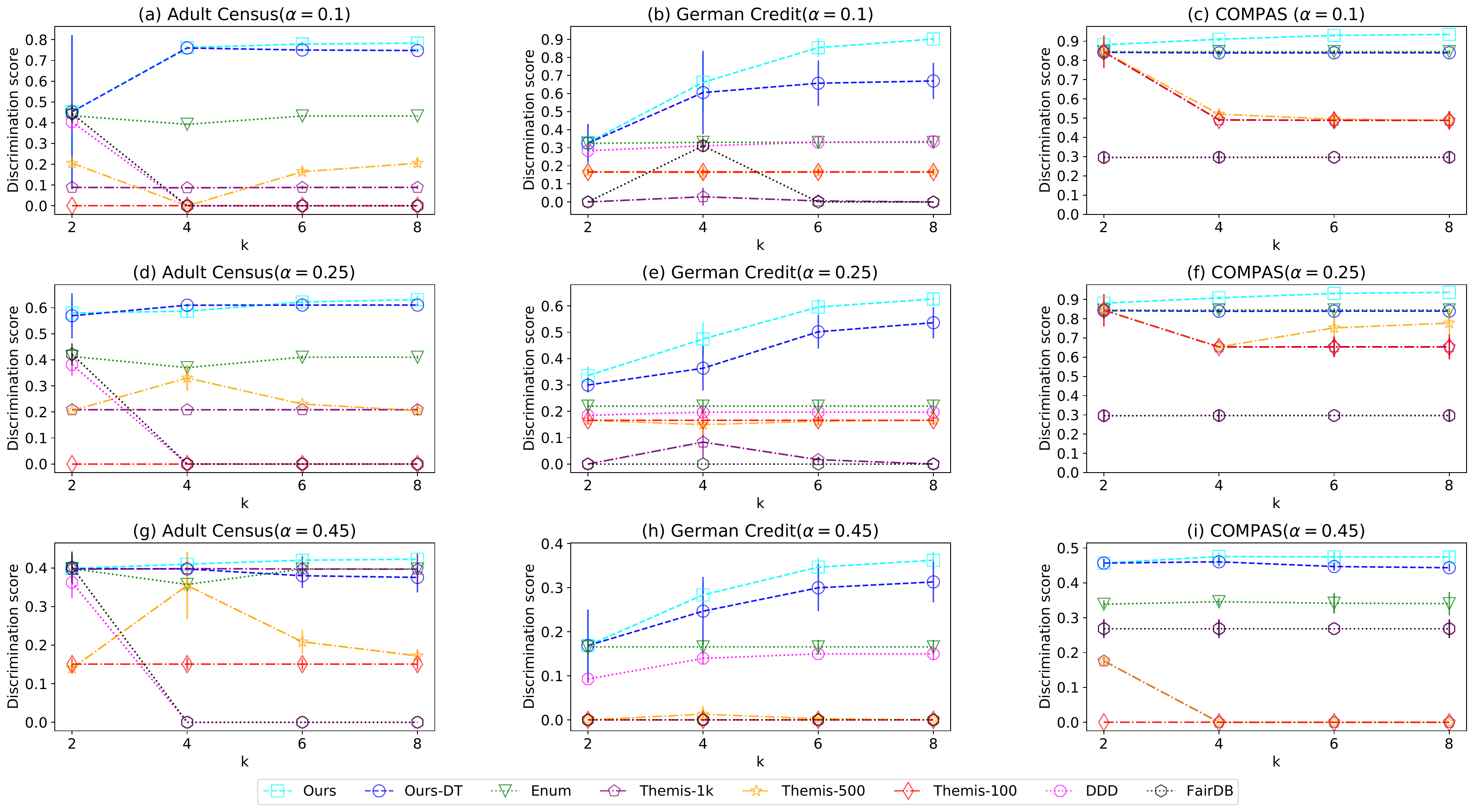}

    \end{center}
\caption{The discrimination score of all compared methods on each of the data sets.}
\label{fig:solution}
\end{figure*}

\begin{figure*}[h!]
    \begin{center}
        \includegraphics[width=1.0\linewidth]{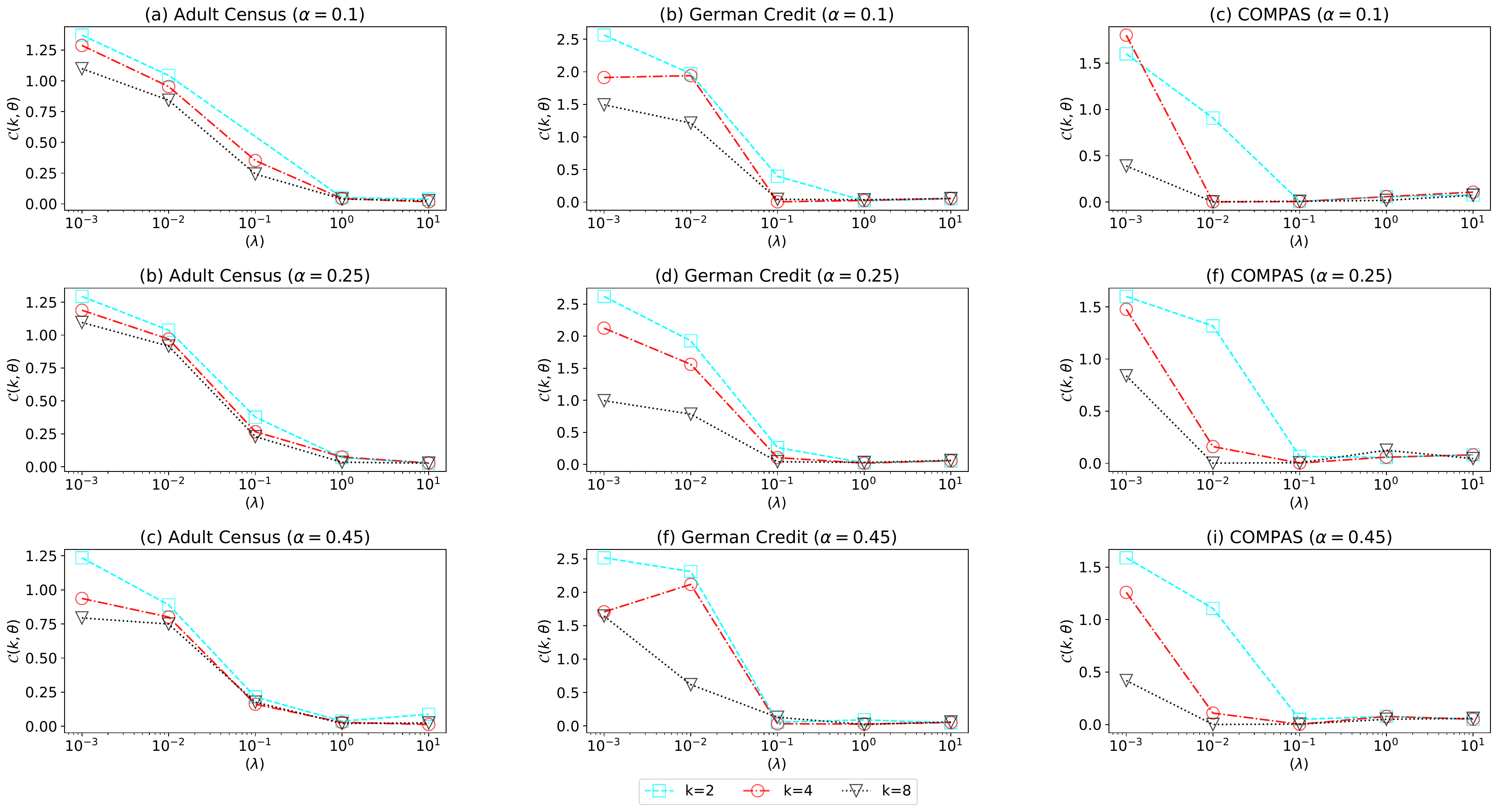}
    \end{center}
\caption{The effect of $\lambda$ on the term $\mathcal{C}(k, \boldsymbol\theta)$.}
\label{fig:lambda}
\end{figure*}

Figure~\ref{fig:case_study}(b) shows another case study on the German Credit data set.
IE-DT successfully finds a discriminated group $\mathcal{S}$ with $DScore(\mathcal{S})=0.22$. 
The decision tree further provides a clear understanding on who the people in $\mathcal{S}$ are and why they are discriminated.
Take the sub-group $G_1$ as an example, a person who is not male has a probability of 0.67 to receive a low predicted risk. Comparing to the males in $G_4$ who has a probability of 0.87 to receive a low predicted risk, the not-male people in $G_1$ are discriminated by the model based on sex.  

A closer look at the decision tree finds that the model may also be unfair within the group of males. 
The males in $G_4$ has a much higher probability to receive a low credit risk than the other males in $G_2$ and $G_3$, who do not own a house or own a house but do not have a short duration of loan.
This reveals the heavy bias of the model towards producing more favorable predictions for the males who own a house and have a shorter duration of loan.

Figure~\ref{fig:case_study}(c) shows the case study on the COMPAS data set, where IE-DT successfully finds a discriminated group $\mathcal{S}$ with $DScore(S)=0.45$. 
We can see from the decision tree that, when compared with the rest of prisoners, the predictive model is heavily biased to offer much more favorable predictions to prisoners with no more than 3 prior offences and are older than 25.
This is solid evidence to reveal the unfairness of the predictive model COMPAS. The interpretable insights from the decision tree on how age and prior offences affect the prediction results are quite consistent with the analysis reported by Propublica~\cite{propub}.

\subsection{Performance Evaluation}
\label{sec:performance}
In this section, we quantitatively evaluate all the methods based on their scalability and quality of their solution. To achieve this, we rely on the metrics of runtime and discrimination score.
For all the methods, the performance in runtime and discrimination score is affected by the thresholds $\alpha$ and $\beta$ on the size of the discriminated group, and the number of key attributes, denoted by $k=|\mathcal{Q}|$.
For the fairness of experiments, we compare the performance of all methods for the same values of $\alpha$, $\beta$, and $k$, where $\alpha\in\{0.1, 0.25, 0.45\}$, $\beta=1-\alpha$, and $k\in\{2, 4, 6, 8\}$.
For IE and IE-DT, we use $\lambda=1$ in Equation~\eqref{eq:ContOptim} for all the experiments.

For each data set, we produce 10 sampled data sets with the same size as the original data set by sampling data instances with replacement from the original data set. We evaluate the performance of every method on each of the 10 sampled data sets and report the mean and standard deviation of the performance.



Figure~\ref{fig:runtime} shows the runtime of every method on each of the data sets. We plot runtime of all of the methods on y-axis while $k$ on x-axis. As $k$ increases, there is exponential increase in the number of possible combinations of attribute assignments to be considered for finding the most discriminated group. Hence, this set-up allows us to comment on the scalability of the considered methods. Due to the high time complexities of Enum, Themis, DDD and FairDB, they are not able to terminate in practical amount of time.
Therefore, we set-up a time budget for all methods to be 5,000 seconds. When a method reaches the time budget, we stop it and return the best solution found so far. Since DDD and FairDB rely on available optimized libraries that do not provide intermediate results, we are unable to obtain any solution for such cases when the budget is completely exhausted. 

The runtime of IE and IE-DT remains fairly consistent with increase in $k$ across all of the data sets. This demonstrates the outstanding scalability of IE and IE-DT which is result of continuous optimization proposed in Equation~\eqref{eq:ContOptim}.

Figure~\ref{fig:runtime}(c) shows that Themis is typically the slowest of all the baselines because of the repeated sampling from uniform distribution for all of the possible attribute value combinations of size up to $k$. Thus, the runtime of Themis becomes too high as the value of $k$ increases. Enum's runtime also increases exponentially with increase in value of $k$. The only case where Enum costs less runtime than IE and IE-DT is for COMPAS data set when $k=2$. This is because the COMPAS data set has only 10 attributes per instance thus Enum can efficiently search all the combinations of attribute values when $k=2$.
For the other cases in Figure~\ref{fig:runtime}, the runtime of Enum is much larger than IE and IE-DT because the time cost of brute-force search grows exponentially with respect to $k$.
FairDB achieves good runtime on COMPAS data set for all $k$ and on the other two data sets for $k=2$. However, as $k$ approaches $8$ for German Credit and Adult Census data sets, the runtime of this approach explodes and it becomes impractical. This is due to huge increase in Approximate Conditional Functional Dependencies possible with increase in value of $k$. Therefore, this approach is also not scalable.
Similarly, DDD achieves great runtime for COMPAS and German Credit data sets but as $k$ increases for Adult Census dataset, its runtime also increases exponentially due to large increase in possible itemsets, and therefore making it unscalable. 

Another comparison that we can draw from Figure~\ref{fig:runtime} is that how runtime changes for the methods when they are run on COMPAS data set that has mere 10 attributes versus the runtime for Adult Census data set with 100 attributes. We can clearly observe that runtime for IE and IE-DT increase by a significantly lower value as compared to other methods. For instance, all of the methods except IE and IE-DT exhaust their time budget on Adult Census data set for $k=4,8$, where as most of them are quite efficient for COMPAS dataset. This is another solid proof of our method's excellent scalability. 

Figure~\ref{fig:solution} shows the discrimination score of the solutions found by all compared methods and how they change with $k$.
We can see that IE and IE-DT achieve the best discrimination score in most of the cases, which demonstrates their outstanding performance in finding solid evidence to reveal the unfairness of a model.
The discrimination score of IE-DT is slightly inferior to that of IE because IE-DT trades some discrimination score for better interpretability. Therefore, enhanced interpretabilty comes at a price of small drop in the discrimination score. The outstanding interpretability of IE-DT has been well demonstrated by the case studies in Section~\ref{sec:case}.

The discrimination scores for the groups found by Enum baseline are significantly lower than those found by IE and IE-DT in most of the cases.
This is because Enum exhausts the set time budget and therefore only returns sub-optimal solution. Also Enum may rely on different decision boundaries than those used by IE, IE-DT to separate $\mathcal{S}$ from $\mathcal{D}\setminus\mathcal{S}$. Recall that our method is able to find discriminating patterns by using a linear decision boundary induced by $f_{\boldsymbol\theta}$. This provides our method with the capability to identify a discriminated group $\mathcal{S}$ that may be defined by multiple attribute assignments rather than just a single attribute assignment. Therefore, a discriminated group identified by the decision boundary of our method may not be exactly identified by another method.

The discrimination scores corresponding to the groups identified by Themis are substantially lower than our method. 
As discussed in Section~\ref{sec:related}, this is because the results produced by Themis are based on synthetic data instances that are independent from the real data.

Broadly speaking, the performance of DDD is significantly worse than our method except for few cases. This is because the frequent itemset mining method cannot handle continuous attributes effectively. It relies on quantization that leads to information loss and inferior performance. 
Similarly, FairDB relies on mining of ACFDs and suffers from similar limitations. As evident in Figure~\ref{fig:solution}, it is significantly worse than our method.

\subsection{Hyperparameter analysis}
We now discuss the effect of $\alpha$ and $\lambda$ on our experiments. 
We see from Figure~\ref{fig:solution} that as the value of $\alpha$ increases, the corresponding values of discrimination score tend to decrease for all the approaches. This is expected as with the increase in $\alpha$, the size constraint on groups $\mathcal{S}$ and $\mathcal{D}\setminus\mathcal{S}$ tightens. Both the groups need to be at least $\alpha*|\mathcal{D}|$ in size. Remember that $\beta$ is set to $1-\alpha$ for the experiments. 

From Figure~\ref{fig:runtime}, we also see that runtime of some approaches like DDD and FairDB increase slightly with decrease in $\alpha$. This can be clearly seen for German Credit and COMPAS data sets where the time budget is not fully exhausted for these approaches. This is expected because the search space for DDD and FairDB is adapted to those combinations that meet the minimum support criteria set by $\alpha$~\cite{han2000mining,caruccio2015relaxed}. Themis and Enum do not involve this step but rather filter out combinations by iterating over them, which makes their runtime insensitive to choice of $\alpha$. Our method relies on continuous optimization that we solve using fixed number of iterations of backpropagation, and therefore runtime is not expected to be affected by the choice of $\alpha$.  

We present the Figure~\ref{fig:lambda} to show the effect of $\lambda$ on the term $\mathcal{C}(k, \boldsymbol\theta)$ which corresponds to the mean of the absolute values of all of the entries in $\boldsymbol\theta$ except the top-$k$.

 We see that as the value of $\lambda$ increases to one, the term $\mathcal{C}(k, \boldsymbol\theta)$ tends to zero for all of the cases. This indicates that all of the entries of $\boldsymbol\theta$ other than the top-k entries are pushed to zero. This would ensure that the maximum number of sensitive attributes considered for uncovering the discrimination are limited to $k$. Therefore, we choose $\lambda=1$ for all of our experiments.

\subsection{Implementation details}

We use SGD optimizer~\cite{robbins1951stochastic} for solving the continuous optimization formulated in Equation~\eqref{eq:ContOptim}. We set learning rate to 0.1 and number of iterations to 2500. For stable learning, we clip the gradients if the norm is greater than 5. All of the experiments are conducted on a Linux machine with an Intel i7-8700K processor and a
Nvidia GeForce GTX 1080 Ti GPU with 11GB memory. Our code is implemented using python 3.7.9 with Pytorch 1.7.1 that uses CUDA version 10.2. For better reproducibility, we make our code available at \cite{our_code}.

%% file: sections/7_Conclusion.tex
\section{Conclusion}
In this paper, we propose the novel task of revealing unfair models by identifying the most discriminated group that is characterized by a small subset of sensitive attributes and their corresponding assignments. 
We present a novel probabilistic framework to model this as a continuous optimization
problem, which can be efficiently solved by a gradient-based
optimization method. Solving the continuous optimization can efficiently and effectively find solid evidence in data to reveal the unfairness of a target model. To enhance interpretability of the evidence, we propose an additional mechanism to clearly explain the evidence in the form of a compact decision tree. Case studies and extensive experiments presented in the paper verify the effectiveness and the scalability of our approach in generating highly interpretable evidence for revealing the discrimination.